\documentclass{article}

\usepackage{PRIMEarxiv}

\usepackage[utf8]{inputenc} 
\usepackage[T1]{fontenc}    
\usepackage{hyperref}       
\usepackage{url}            
\usepackage{booktabs}       
\usepackage{amsfonts}       
\usepackage{nicefrac}       
\usepackage{microtype}      
\usepackage{lipsum}
\usepackage{fancyhdr}       
\usepackage{graphicx}       
\graphicspath{{media/}}     
\usepackage[utf8]{inputenc} 
\usepackage[T1]{fontenc}    
\usepackage{hyperref}       
\usepackage{url}            
\usepackage{booktabs}       
\usepackage{amsfonts}       
\usepackage{nicefrac}       
\usepackage{microtype}      
\usepackage{graphicx}
\graphicspath{ {./images/} }
\usepackage{xcolor}         
\usepackage{amsmath}        
\usepackage{algorithm}
\usepackage{algpseudocode}
\usepackage{float}
\usepackage[accsupp]{axessibility}  
\usepackage[T1]{fontenc}
\usepackage{graphicx}
\usepackage{booktabs}
\usepackage{multirow}
\usepackage{algorithm}
\usepackage{algpseudocode}
\usepackage{float}
\usepackage{subcaption}
\usepackage{comment}
\usepackage{ulem}
\usepackage{xcolor}
\usepackage{amsmath}
\usepackage{tcolorbox}
\usepackage{color}
\usepackage{soul}
\sethlcolor{yellow}
\usepackage{breqn}
\usepackage{colortbl}
\usepackage{booktabs} 
\usepackage{array}

\title{TabSeq: A Framework for Deep Learning on Tabular Data via Sequential Ordering\thanks{This paper has been accepted for presentation at the 27th International Conference on Pattern Recognition (ICPR 2024) in Kolkata, India.}}

\author{
  Al Zadid Sultan Bin Habib$^{1}$, 
  Kesheng Wang$^{2}$, 
  Mary-Anne Hartley$^{3}$, 
  Gianfranco Doretto$^{4}$, 
  Donald A. Adjeroh$^{5}$ \\
  \\
  $^{1,4,5}$Lane Department of Computer Science and Electrical Engineering, West Virginia University, Morgantown, WV 26505, USA \\
  \texttt{ah00069@mix.wvu.edu\textsuperscript{1}}, \texttt{gianfranco.doretto@mail.wvu.edu\textsuperscript{4}}, \texttt{donald.adjeroh@mail.wvu.edu\textsuperscript{5}} \\
  $^{2}$College of Nursing, University of South Carolina, Columbia, SC 29208, USA \\
  \texttt{kesheng@mailbox.sc.edu\textsuperscript{2}} \\
  $^{3}$Yale University School of Medicine, New Haven, CT 06510, USA \\
  \texttt{mary-anne.hartley@yale.edu\textsuperscript{3}} \\
} 

\begin{document}
\maketitle

\begin{abstract}
Effective analysis of tabular data still poses a significant problem in deep learning, mainly because features in tabular datasets are often heterogeneous and have different levels of relevance. This work introduces TabSeq, a novel framework for the sequential ordering of features, addressing the vital necessity to optimize the learning process. Features are only sometimes equally informative, and for certain deep learning models, their random arrangement can hinder the model’s learning capacity. Finding the optimum sequence order for such features could improve the deep learning models’ learning process. The novel feature ordering technique, which we provide in this work, is based on clustering and incorporates both local ordering and global ordering. It is designed to be used with a multi-head attention mechanism in a denoising autoencoder network. Our framework uses clustering to align comparable features and improve data organization. Multi-head attention focuses on essential characteristics, whereas denoising autoencoder highlights important aspects by rebuilding from distorted inputs. This method improves the capability to learn from tabular data while lowering redundancy. Our research demonstrating improved performance through appropriate feature sequence rearrangement utilizing raw antibody microarray and two other real-world biomedical datasets validates the impact of feature ordering. These results demonstrate that feature ordering can be a viable approach to improved deep learning of tabular data.
\end{abstract}


\section{Introduction}
\vspace{-0.25cm}
\label{sec:intro}
Deep learning has transformed how we handle and comprehend diverse data types, resulting in unparalleled progress in numerous fields. Deep learning models have outperformed conventional techniques in audio analysis, picture identification, and Natural Language Processing (NLP), opening the door to new applications previously thought impractical. For example, Convolutional Neural Networks (CNNs) have emerged as the mainstay of image-processing applications, demonstrating exceptional performance in picture classification, object recognition, and other applications \cite{b48}, \cite{b49}. In NLP, Transformer-like models have established new benchmarks for text summarization, machine translation, and question-answering systems \cite{b38}, \cite{b50}. Additionally, deep learning has helped audio processing by advancing speech recognition and synthesis, greatly enhancing user interaction with technology \cite{b51}, \cite{b52}, \cite{b53}. These achievements demonstrate how deep learning is an essential tool for applications where standard feature engineering fails because it can grasp intricate patterns and relationships inside high-dimensional data.

The quest for an optimal deep learning architecture for tabular data, crucial in sectors like finance, healthcare, and retail, remains ongoing. Unlike image, text, and audio data, tabular data's structure, rows representing samples, and columns as features present distinct challenges, especially in modeling complex feature relationships that lack spatial or sequential correlation. Innovative model architectures and data representation methods are essential to address tabular data's unique aspects. Models such as TabNet\cite{b1}, Neural Oblivious Decision Ensembles (NODE)\cite{b6}, and TabTransformer\cite{b2} have emerged as practical solutions alongside popular gradient-boosting tree models. However, gaps remain in handling scenarios with high-dimensional features against smaller sample sizes, such as genomic or other medical data.

We introduce TabSeq, a framework for deep learning on tabular data, using the feature ordering to optimize tabular data utilization.  Our approach is motivated by methods of band ordering often used in the efficient analysis of hyperspectral images.
Adapting band ordering from hyperspectral images\cite{b54} to tabular data involves comparing dataset features to spectral bands, where features, like bands, vary in informational value. This approach uses statistical and machine learning methods to prioritize significant features and reduce redundancy, enhancing dataset efficiency similar to compression in hyperspectral imaging. The bandwidth minimization problem in communication networks\cite{b55} focuses on optimizing data transmission order or compressing data to meet bandwidth limits, akin to arranging features in tabular data for deep learning models. The novel contributions of our paper are as follows:
\vspace{-0.25cm}
\begin{enumerate}
    \item We present a novel feature ordering technique that combines local ordering and global ordering to optimize feature sequences and clustering to group comparable features. This innovative method systematically improves learning and significantly improves the model's performance on tabular datasets by prioritizing features according to their relevance and informative content.
    \item Our framework enables a Denoising Autoencoder (DAE) architecture to incorporate the Multi-Head Attention (MHA) mechanism smoothly. This integration highlights important characteristics, eliminates redundancy by rebuilding inputs from partially corrupted versions, and allows for dynamic attention to vital elements.
    \item Our studies using raw antibody microarray and other datasets show that our feature ordering approach substantially improves the performance of deep learning models. The outcomes demonstrate how feature sequencing is crucial for training and validating the potential of feature ordering in tabular data processing inside deep learning frameworks.
\end{enumerate}
These contributions collectively address the challenges of heterogeneous feature relevance in tabular data, setting a new precedent for data preprocessing and model optimization in deep learning applications.
\vspace{-0.3cm}
\section{Related Work}
\vspace{-0.2cm}
Feature ordering in tabular datasets is essential for improving machine learning models' interpretability, accuracy, and efficiency, particularly in deep learning. Models that recognize and rank important features can learn new information more quickly, require less training time, and exhibit better generalization on unobserved data \cite{b67}. While feature ordering is essential for all tabular data types, including numerical data, it influences models that use data structures, such as attention mechanisms or specific autoencoders.

\textbf{Attention-based Models:} TabNet\cite{b1} employs an attention mechanism for feature selection in tabular data, enhancing performance and interpretability without rearranging features. TabTransformer\cite{b2} uses contextual embeddings to improve accuracy in handling categorical data, though it requires pre-training and fine-tuning. AutoInt\cite{b3} specializes in Click-Through Rate (CTR) prediction by learning feature interactions with a self-attentive network despite assuming unordered features. ASENN\cite{b4} predicts pavement deterioration with multi-dropout attention layers, offering efficient infrastructure maintenance solutions. Attention-based models might find it difficult to determine how important a particular feature is to the model's predictions; feature ordering can help with this problem by highlighting the elements with the most significant impact.

\textbf{Tree-based Models:} The Tree Ensemble Layer (TEL)\cite{b5} by Hazimeh et al. enhances neural networks with the efficiency of tree ensembles through ``soft trees'' and sparse activations, improving performance. TEL, however, does not perform well in capturing complex feature interactions. NODE\cite{b6} by Popov et al. combined deep learning flexibility with gradient-boosted with the benefit of decision trees. They achieved superior outcomes via differentiable trees and entmax transformation, albeit with potential limitations in capturing nuanced feature interactions. Tree-based models might struggle to explain complex feature associations; feature ordering fills this gap by arranging features to clarify their relationships.

\textbf{LLM-based Models:} TabLLM\cite{b7}, developed by Hegselmann et al., leverages LLMs for few-shot categorization of tabular data by translating tables into natural language, showing superior performance over traditional techniques with limited data. MediTab\cite{b8} by Wang et al. introduced a ``learn, annotate, refine'' approach combined with LLMs for medical data predictions, achieving high performance and excellent zero-shot capabilities without fine-tuning. IngesTables\cite{b9}, presented by Yak et al., creates scalable tabular foundation models, addressing key issues, such as large cardinality and semantic relevance, through an attention-based method with LLMs, offering cost-effective alternatives to conventional models for clinical trial predictions. Implicit data hierarchies may be a problem for LLM-based models; feature ordering might help by arranging data to represent underlying importance and relationships.

\textbf{Graph-based Models:} Ruiz et al. introduced PLATO\cite{b10}, leveraging an auxiliary knowledge graph for model regularization in MLPs. This improved learning on high-dimensional tabular datasets, reducing over-fitting by connecting features to knowledge graph nodes. T2G-FORMER\cite{b11}, by Yan et al., enhances structured feature interactions through a Graph Estimator and a Transformer network, offering superior interaction modeling and prediction accuracy over conventional deep neural networks. Chen et al. proposed HyTrel\cite{b12}, a model using hypergraphs to capture tabular data's structural properties, outperforming existing methods with minimal pretraining by integrating inductive biases about data structure. Graph-based models may miss linear feature correlations; feature ordering can address this limitation by better aligning features to depict linear trends and relationships. In general, graph-based approaches provide improvements in tabular data analysis by capturing possible interactions between features. However, they suffer from high computational costs, limiting their applicability.

\textbf{Autoencoder-based Models:} ReConTab, developed by Chen et al., is a deep learning framework for automatic representation learning from tabular data, utilizing contrastive learning and an asymmetric autoencoder with regularization to enhance classification models like Random Forest and XGBoost\cite{b13}. ReMasker, introduced by Du et al., employs masked autoencoding for imputing missing values in tabular data. This improved the results through randomization in masking extra values and offering competitive performance, especially with increased missing data\cite{b14}. SwitchTab offers a self-supervised learning approach to identify less apparent dependencies in tabular data, using an asymmetric encoder-decoder to improve prediction tasks and provide interpretable insights via its embeddings\cite{b15}. Autoencoder-based models may need to be more efficient in prioritizing influential features, a limitation that feature ordering can overcome by arranging features to enhance model focus and interpretability.

\textbf{Other Models:} GrowNet introduces a method leveraging shallow neural networks within a gradient-boosting framework for various machine learning tasks, limited by static feature selection \cite{b16}. Using Scaled Exponential Linear Units (SELU), Self-Normalizing Neural Networks (SNNN) aim for automatic activation normalization to stabilize deep learning but face restrictions due to reliance on SELUs \cite{b17}. DCN V2 advances the integration of feature interactions, constrained by its static interaction framework \cite{b18}. Gorishniy et al.'s critique of deep learning models for tabular data, including the FT-Transformer, highlights the need for dynamic feature sequencing to enhance model performance \cite{b19}. Static feature integration is a potential problem for these models, which feature ordering can solve by dynamically modifying feature sequences to maximize learning and performance. Also see the transformer-based model, TabPFN, and the regularization-based model, TANGOS, in \cite{b69} and \cite{b68} respectively.
\newline

The literature review highlights innovative strategies for enhancing deep learning model performance on tabular data. Our feature ordering approach uniquely merges clustering with local ordering and global ordering in an MHA-augmented DAE framework, focusing on the ordered arrangement of features based on their importance and information content. This strategy contrasts with traditional methods that often do not consider the arrangement of features. 
\vspace{-0.4cm}
\section{Methodology}
\begin{figure}
    \centering
    \includegraphics[width=0.7\linewidth]{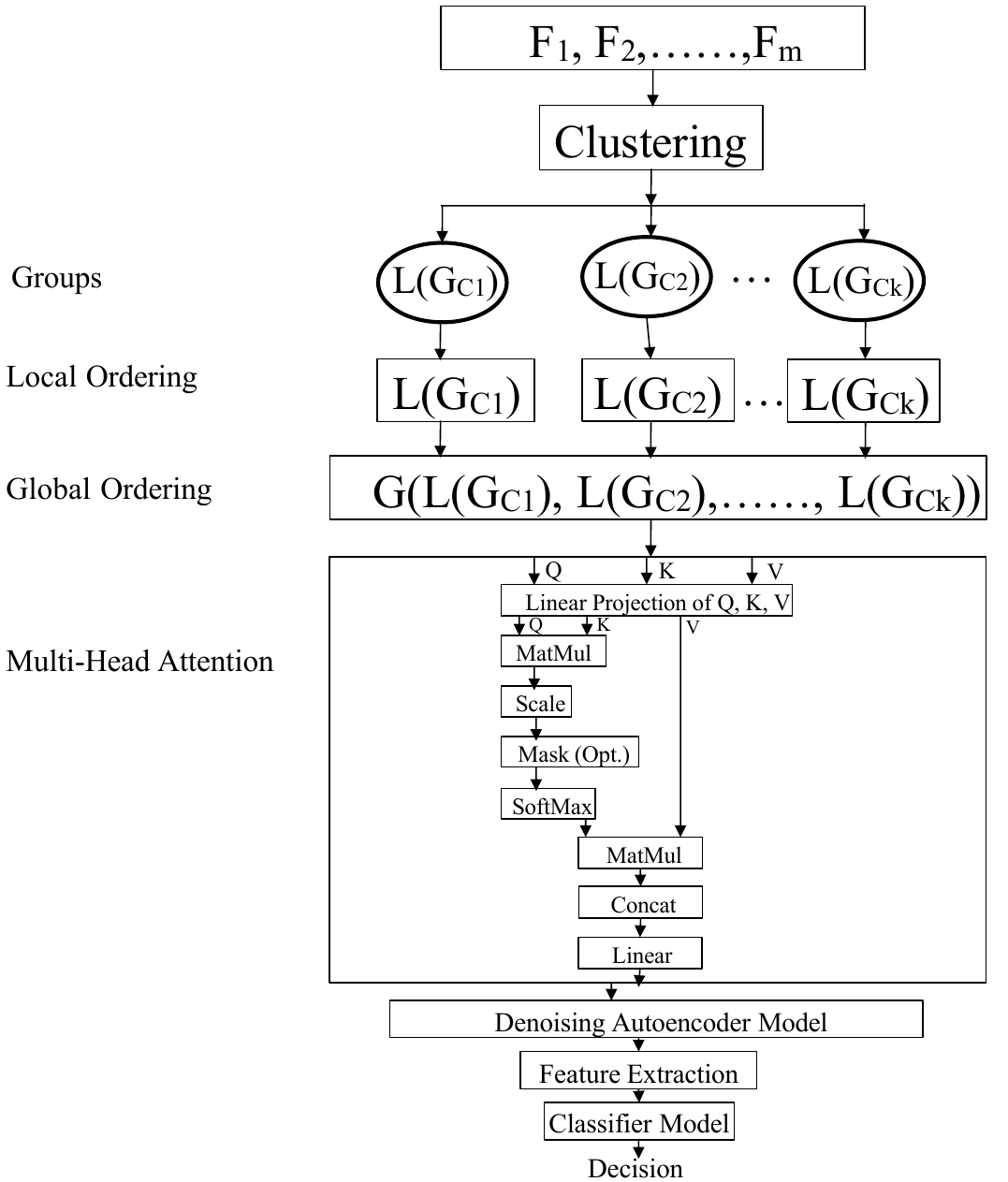}
    \caption{Overview of the TabSeq framework.}
    \label{fig:1}
\end{figure}

\vspace{-0.4cm}
This section presents our deep learning architecture, designed primarily to analyze tabular data effectively, as seen in Figure \ref{fig:1}. The process begins with feature clustering and then moves to local ordering and global ordering to improve input feature arrangements. These rearranged features are fed into an MHA mechanism and a DAE, ultimately leading to feature extraction and classification model decisions. Our methodology introduces a novel feature ordering technique to improve the analysis of tabular datasets. This methodology seeks to improve prediction accuracy and robustness for various applications by capturing intricate feature interactions and underlying patterns in the data.
\vspace{-0.4cm}
\subsection{Feature Ordering}
\vspace{-0.4cm}
Feature ordering is finding an optimal arrangement of features within and across clusters to minimize a defined cost function that reflects the disorganization of feature positioning. This involves computing permutations that best sequence the features according to their relationships. Given a dataset \( X \in \mathbb{R}^{n \times m} \) with \( n \) samples and \( m \) features, we define a set of graphs \( \{G_1, G_2, ..., G_k\} \), where \( k \) is the number of clusters, and each graph \( G_c = (V_c, E_c) \) for cluster \( c \) has vertices \( v_i \in V_c \) corresponding to features within that cluster. The edges \( (v_i, v_j) \in E_c \) represent significant relationships between features \( i \) and \( j \) within the cluster. For each cluster \( c \), the goal is to find a permutation \( \pi_c \) of its features that minimizes a local cost function \( F_c \):
\[ F_c(\pi_c) = \sum_{(v_i, v_j) \in E_c} |i - j| \]
where \( i = \pi_c(v_i) \) and \( j = \pi_c(v_j) \) represent the indices of features \( v_i \) and \( v_j \) in the permutation \( \pi_c \), minimizing the disorganization within the cluster. The overall goal is to find a global permutation \( \Pi \) that integrates the local permutations \( \pi_c \)'s and minimizes a global cost function \( F_G \):
\[ F_G(\Pi) = \sum_{c=1}^{k} \alpha_c \cdot F_c(\pi_c) \]
where \( \alpha_c \) represents the weight or importance of cluster \( c \) in the global context with $\sum_{c} \alpha_c =1$. The optimal permutation \( \Pi^* \) minimizes \( F_G(\Pi) \):
\[ \Pi^* = \arg\min_{\Pi} F_G(\Pi) \]

\textbf{Local Ordering Function:} Local ordering is the computation of a permutation that minimizes the sum of absolute differences in positions of related features within a cluster, thereby reducing feature dispersion.
\[ L(G_c) = \text{argmin}_{\pi_c} \sum_{(v_i, v_j) \in E_c} | \pi_c(i) - \pi_c(j) | \]
Here, \( L(G_c) \) outputs the permutation \( \pi_c \) for cluster \( c \) that minimizes the feature dispersion, where \( \pi_c(i) \) is the position of feature \( i \) in the permutation \( \pi_c \), and the sum quantifies the total dispersion of features in the cluster.

\textbf{Global Ordering Function:} Global ordering is the process of integrating local permutations from all clusters into a global permutation that minimizes the weighted sum of within-cluster feature dispersion to enhance the deep learning model's performance.
\[ G(\{\pi_1, \pi_2, ..., \pi_k\}) = \text{argmin}_{\Pi} \sum_{c=1}^{k} \alpha_c \cdot D_{\pi_c}(G_c) \]
\( G \) integrates the local permutations \( \pi_c \) of all clusters into a global ordering \( \Pi \), minimizing the weighted sum of feature dispersion \( D_{\pi_c}(G_c) \) within each cluster \( G_c \), where \( \alpha_c \) represents the weight or importance of cluster \( c \). In these functions, \( D_{\pi_c}(G_c) \) represents a measure of feature dispersion within cluster \( c \) based on the permutation \( \pi_c \), and \( \Pi \) is the global permutation that integrates these local orderings into a dataset-wide global feature ordering that aims to improve the performance of the model.

\textbf{Feature Dispersion:} 
In the context of feature ordering for a tabular dataset, the term ``Feature Dispersion'' describes the degree to which features with a strong relationship (or dependency) are placed far apart in the ordering. The goal would be to minimize this dispersion so that related features are positioned closer together, which could be advantageous for specific deep learning models that can benefit from the structure of the data (See statistical dispersion in \cite{b62}, \cite{b63}).

For instance, a generalized feature dispersion for a cluster \( G_c \) could be defined as:
\[ D(\pi_c) = \sum_{(v_i, v_j) \in E_c} w_{ij} \cdot | \pi_c(i) - \pi_c(j) | \]
Where,
\( \pi_c \) is the permutation of features within cluster \( c \),
\( w_{ij} \) is a weight assigned to the edge between features \( i \) and \( j \) (which could be based on the strength of the relationship between the features e.g., correlation or mutual information),
\( |\pi_c(i) - \pi_c(j)| \) is the absolute difference in the ordered positions of features \( i \) and \( j \) within the permutation \( \pi_c \).

\textbf{Feature Dispersion and Variance:}
We adopt variance as a metric to guide the ordering of features locally \cite{b66}. This approach is based on the premise that organizing features to minimize their dispersion within clusters can enhance model performance by affecting the variance of these features in a beneficial manner. We understand that feature dispersion within a cluster, \(D(\pi_c)\), reflects how spread out the features are in terms of their arrangement or ordering based on certain criteria (e.g., importance, similarity, etc.). Variance, \(\text{Var}(X_i)\), measures the spread of values for a given feature \(i\) across the dataset or within clusters. The goal is to understand how minimizing \(D(\pi_c)\) influences \(\text{Var}_c(X_i)\) for features within the same cluster. A decrease in \(D(\pi_c)\) (i.e., reduced dispersion or more closely arranged features based on their relationships) leads to an increase in the homogeneity of feature values within the cluster. This homogeneity, in turn, can lead to a more meaningful and possibly reduced variance (\(\text{Var}_c(X_i)\)) for the features within the cluster, as related features that behave similarly or have strong relationships are positioned closer together, thus reflecting their actual data distribution more accurately. The conceptual relationship can be summarized as:
\[ \text{Var}_c(X_i) \propto \frac{1}{D(\pi_c)} \]
This expression suggests that as feature dispersion within a cluster decreases (making \(D(\pi_c)\) smaller), the variance of features within that cluster (\(\text{Var}_c(X_i)\)) becomes more meaningful of the true data distribution. The inverse proportionality indicates that lower dispersion (closer grouping of related features) leads to a more stable or accurate variance representation, underlining the importance of thoughtful feature arrangement in enhancing model understanding and performance. Algorithm 1 captures the general procedure for our feature ordering, including both steps of local and global ordering.

\begin{algorithm}
\caption{Feature Ordering}
\begin{itemize}
    \item Preprocessed dataset \(X_{\text{train}} \in \mathbb{R}^{n \times m}\) with \(n\) samples and \(m\) features.
    \item Clustering Algorithm: Choose from k-means, DBSCAN, HDBSCAN, or a Custom Algorithm.
    \item Sorting Order: Ascending or Descending.
    \item Number of clusters (num\_clusters), required for k-means and optional for other algorithms.
\end{itemize}

\textbf{Output:}
\begin{itemize}
    \item Reordered dataset \(new\_training\_set \in \mathbb{R}^{n \times m}\).
\end{itemize}

\textbf{Procedure}
\begin{algorithmic}[1]
    \State Initialize the clustering model based on the selected algorithm.
    \If{Clustering Algorithm is k-means}
        \State Specify num\_clusters.
    \Else
        \State Use default or custom settings.
    \EndIf
    \State Apply clustering to \(X_{\text{train}}\) to get cluster labels.
    \State Append cluster labels to \(X_{\text{train}}\).
    \For{each cluster, excluding noise}
        \State Select data for the current cluster.
        \State Calculate and order feature dispersion based on the sorting order.
        \State Record feature order for the cluster.
    \EndFor
    \State Combine feature orders from all clusters into an overall feature order.
    \State Reorder \(X_{\text{train}}\) columns according to the overall feature order.
    \State Assess model performance with the reordered dataset.
\end{algorithmic}
\end{algorithm}
\vspace{-0.2cm}
\subsection{MHA}
MHA or Multi-Head Attention, inspired from\cite{b38}, is the integration that serves as a cornerstone for enhancing the model's capacity to capture complex dependencies and interactions within the input data. This mechanism's key feature is its capacity to narrow down an input sequence through several attention heads at once, which enables the model to pay attention to data from various representation subspaces at various points in time. Formally, for each head $h$, we perform linear transformations on the input $X$ to obtain queries $Q_h$, keys $K_h$, and values $V_h$ using parameter matrices $W^Q_h$, $W^K_h$, and $W^V_h$, respectively:
$$
Q_h = XW^Q_h, \quad K_h = XW^K_h, \quad V_h = XW^V_h
$$
Subsequently, we compute the scaled dot-product attention for each head. The attention function operates on queries, keys, and values and scales the dot products of queries with all keys by $\frac{1}{\sqrt{d_k}}$, where $d_k$ is the dimensionality of the keys and queries. This scaling factor helps stabilize the gradients during training. The attention scores are then passed through a softmax function to obtain the weights on the values:
$$
\text{Attention}(Q_h, K_h, V_h) = \text{softmax}\left(\frac{Q_hK_h^T}{\sqrt{d_k}}\right)V_h
$$
The final output of the MHA layer is created by concatenating and linearly transforming each head's output, each of which captures a unique feature of the incoming data. The information acquired by each head is combined by this concatenation procedure, maintaining the diversity of the attended features:
$$
\text{MultiHead}(Q, K, V) = \text{Concat}(\text{head}_1, ..., \text{head}_h)W^O
$$
where, $\text{head}_h = \text{Attention}(Q_h, K_h, V_h)$, and $W^O$ is the parameter matrix for the output linear transformation. Using this sophisticated attention mechanism, our model gains a more sophisticated capacity to identify and use the complex patterns in the data. The MHA mechanism provides insights into the diverse elements of the data that different heads focus on, hence improving the interpretability and expressive capacity of the model.
\vspace{-0.45cm}
\subsection{DAE}
\vspace{-0.25cm}
TabSeq leveraged an MHA layer in addition to DAE \cite{b41} architecture to overcome the difficulties associated with learning robust representations from high-dimensional tabular data. The DAE enhances the model's performance on ensuing tasks by lowering noise and extracting significant features. The DAE architecture comprises an encoder and a decoder, where the encoder maps the input data \(X\) to a latent space representation \(Z\), and the decoder reconstructs the input from \(Z\). The MHA layer improves the encoder's capacity to focus on relevant information by allocating different levels of attention to different data segments. Formally stated, the encoding procedure is as follows:
\[
Z = f_{\text{encoder}}(X) = \text{ReLU}(W_e \cdot \text{MHA}(X) + b_e)
\]
Where \(X\) is the input data, \(W_e\) and \(b_e\) are the weights and bias of the encoding layer, respectively, and \(\text{ReLU}\) denotes the Rectified Linear Unit activation function. The \(\text{MHA}(X)\) function represents the output of the MHA layer applied to \(X\). The decoder, aiming to reconstruct the input data from the latent representation \(Z\), is given by:
\[
\hat{X} = f_{\text{decoder}}(Z) = \text{Sigmoid}(W_d \cdot Z + b_d)
\]
where, \(\hat{X}\) is the reconstructed data, \(W_d\) and \(b_d\) are the decoder weights and bias, and \(\text{Sigmoid}\) is the activation function facilitating reconstruction. The loss function for the DAE, aiming to minimize the reconstruction error, is defined as the Mean Squared Error (MSE) between the original input \(X\) and its reconstruction \(\hat{X}\):
\[
\mathcal{L}_{\text{MSE}} = \frac{1}{N}\sum_{i=1}^{N}(X_i - \hat{X}_i)^2
\]
where \(N\) is the number of samples in the dataset. The MHA layer added to the DAE architecture improves the encoder's capacity to identify and highlight the most informative characteristics by utilizing the attention mechanism \cite{b42}, \cite{b43}, \cite{b44}, \cite{b45}. This leads to a less noisy and more discriminative representation in the latent space, which is essential for tasks that come after, like classification. Using a sequential network with dense layers, ReLU activation \cite{b46} for hidden layers, and sigmoid activation \cite{b47} for the output layer, optimized for classification tasks, the final classification model is trained on the encoded representations. Our methodology efficiently tackles the problem of learning from high-dimensional and noisy tabular data by incorporating the MHA layer within the DAE framework, significantly improving the model's predictive performance and robustness.
\vspace{-0.45cm}
\subsection{Feature Extraction and Classifier Model}
\vspace{-0.3cm}
In the TabSeq framework, the DAE is instrumental in preprocessing the input data by denoising and enhancing feature salience through its robust feature extraction process, where the encoder component transforms the corrupted input into a refined, lower-dimensional representation. These enhanced features are then utilized by the classifier, which is specifically configured with a softmax activation for multi-class scenarios to generate class probability distributions or a sigmoid activation for binary classification to yield a probability of class membership. This setup ensures that the classifier operates on high-quality features extracted post-DAE, thereby optimizing the model’s accuracy and adaptability to different classification tasks. In a nutshell, the DAE processes the input data, which is then followed by a feature extraction process to extract robust, noise-reduced features, which are then utilized by the classifier to ensure precise predictions based on clean and relevant information, illustrating an essential sequential information flow where the classifier's efficacy is significantly enhanced by the high-quality features provided by the DAE.
\vspace{-0.4cm}
\section{Experimental Results}
\label{results}
\vspace{-0.3cm}
\subsection{Datasets and Model Hyperparameters}
\vspace{-0.2cm}
In our research, the autoimmune diseases dataset used in \cite{b37} and publicly released in \cite{b56} contains 393 features targeting five disease classes of 316 samples, detailing each antibody's signal intensity. The ADNI dataset \cite{b57} includes 177 samples and 263 features with target attributes like AD123, ABETA12, and AV45AB12, representing various stages of Alzheimer's disease and captured through DTI analysis for white matter integrity. Lastly, the WDBC dataset \cite{b61} offers 32 features derived from breast mass images, aiming to differentiate between 357 benign and 212 malignant cases. Each dataset was partitioned into training, validation, and testing subsets following a 70:15:15 split, focusing on specific target attributes for comprehensive classification and analysis. TabSeq model with feature ordering integrates an MHA mechanism with four heads and dimensionality of 32 alongside a DAE comprising dense ReLU-activated layers. It was trained over 50 epochs with a batch size of 32 using the Adam optimizer, and the model employs MSE loss for the DAE and binary cross-entropy loss for the classifier. This configuration facilitates nuanced feature extraction and robust classification, as evidenced by the model's validation performance, optimizing computational efficiency and learning effectiveness. In our analysis, feature ordering was uniformly applied across baseline models using k-means clustering with 5 clusters in ascending order for the autoimmune diseases and ADNI datasets and 3 clusters for the WDBC dataset. This consistent methodology underscores the effectiveness of feature ordering in enhancing model performance across the board, with TabSeq demonstrating particularly notable improvements in accuracy and AUC with feature ordering.  
\begin{figure}
    \centering
    \includegraphics[width=\linewidth]{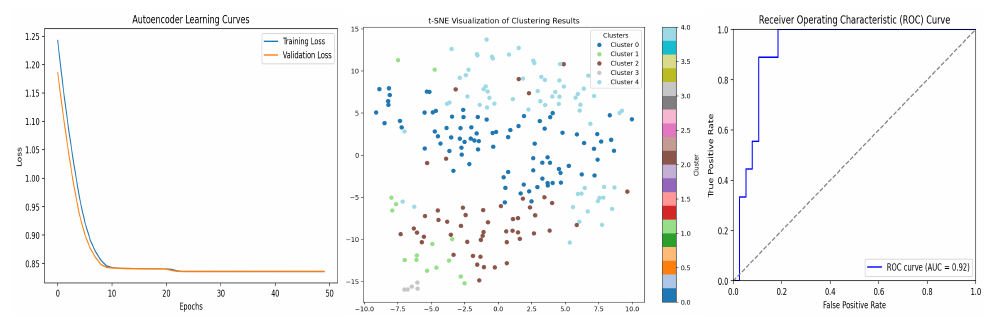}
    \caption{Visualization of model performance on the autoimmune diseases dataset.}
    \label{fig:2}
\end{figure}
\vspace{-0.4cm}
\subsection{Experiments with Autoimmune Diseases' Dataset}
\vspace{-0.2cm}
Two cases were included in our investigation (Table \ref{tab:1}), which showed how feature ordering affected model performance. In Case 1, NODE's accuracy sharply declined, whereas TabSeq's increased from 85.42\% to 87.23\% with feature ordering. Case 2 demonstrated how feature ordering significantly improved TabNet's accuracy, raising it to 94.44\%. TabSeq demonstrated its practical usage of feature ordering with consistently good performance. These findings highlight the context and model-specificity of feature ordering's efficacy, with TabSeq consistently outperforming other models in the dataset on autoimmune disorders. Visualizations in Fig. \ref{fig:2} confirm effective clustering for feature ordering and model generalization and show the ROC curve's high binary classification accuracy. From Table \ref{tab:1}, it is evident that feature ordering significantly affects model performance, especially in sequence-dependent architectures (e.g., autoencoders and LSTM), where the feature ordering aligns features in a meaningful sequence to improve learning. However, tree-based models, such as NODE, could not benefit as much because of their built-in feature selection methods. The local receptive field of 1D CNNs limits the effect of feature ordering. In contrast, TabNet's attention mechanism already prioritizes relevant features, which may cause inconsistent feature ordering performance. TabTransformer consistently enhances AUC across datasets; TANGOS shows marked improvements in AUC with feature ordering but does not demonstrate strong accuracy. These results underline TabSeq's capacity to discern complex patterns in high-dimensional genomic datasets, advocating their potential for data analysis. Based on ablation studies, we chose k-means with 5 clusters for feature ordering with features sorted in ascending order (Table \ref{tab:1} for Case 1-2 and Fig. \ref{fig:2} for Case 1).
\vspace{-0.4cm}
\subsection{Ablation Studies}
\begin{figure}
\centering
\begin{subfigure}{.525\textwidth}
  \centering
  \includegraphics[width=\linewidth]{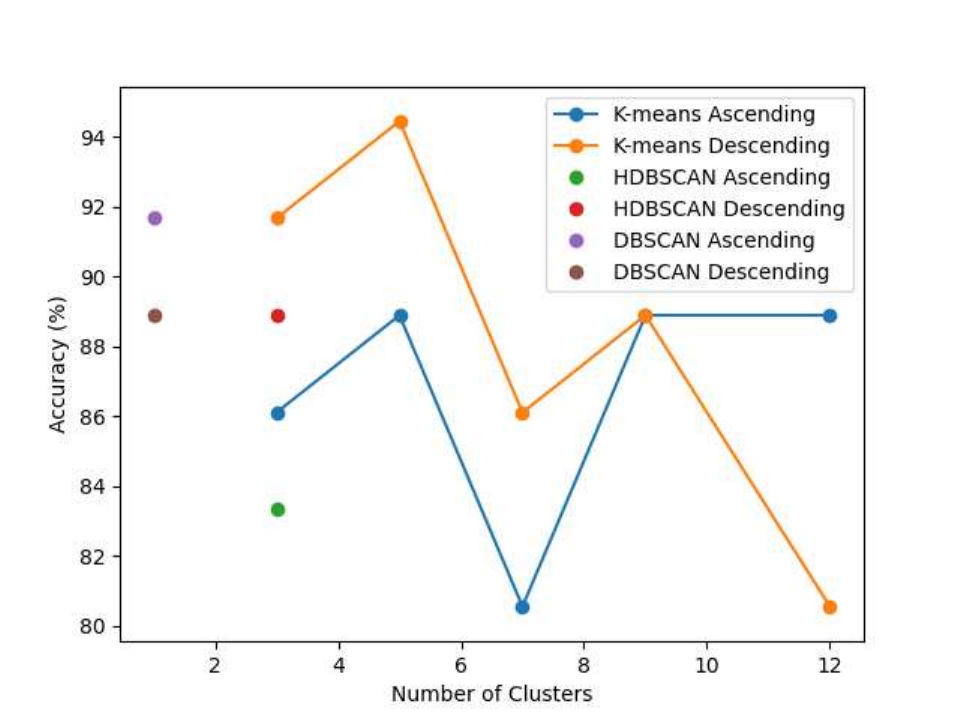}
  \caption{Case 2: Accuracy}
  \label{fig:case2_accuracy}
\end{subfigure}%
\begin{subfigure}{.525\textwidth}
  \centering
  \includegraphics[width=\linewidth]{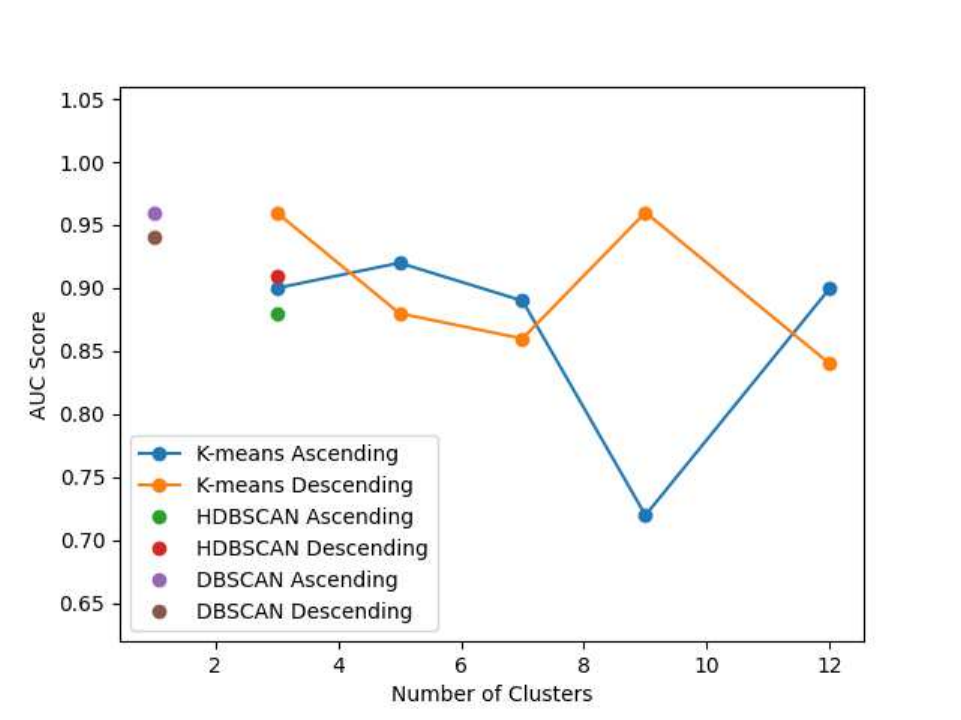}
  \caption{Case 2: AUC}
  \label{fig:case2_auc}
\end{subfigure}
\caption{Results' plots for TabSeq with different feature ordering configurations.}
\label{fig:cases}
\end{figure}

\vspace{-0.3cm}
We assessed the TabSeq model's performance using the autoimmune diseases' dataset, mainly how the clustering algorithms affected feature ordering. Various clustering algorithms affected AUC and model accuracy in different scenarios. For example, in Case 1, the maximum accuracy of 87.23\% and the highest AUC of 0.92 were obtained using DBSCAN with a single cluster with the features sorted in ascending order. Case 2, on the other hand, achieved 94.44\% accuracy and optimal performance using k-means at five clusters with features sorted in descending order. Surprisingly, DBSCAN achieved perfect accuracy and an AUC of 100\% and 1.00 in Case 4, with features sorted in descending order. The significance of the number of clusters, specific clustering algorithm, and sorting order were also observed for these cases. Fig. \ref{fig:cases} shows the assessment for Case 2, whereas the figures for other cases also looked similar. These results highlight the crucial role that clustering configurations in feature ordering play in improving the predictive power of the TabSeq model, indicating that the model's ability to distinguish intricate patterns of autoimmune diseases is greatly influenced by strategic cluster formation and feature ordering.
\subsection{Experiments with ADNI and WDBC Dataset}
The studies done on the ADNI dataset (Table \ref{tab:3ab}) show how effective feature ordering is in several models, such as TabNet, NODE, and TabSeq. Across two target attributes, such as AD123 and ABETA12, the addition of feature ordering resulted in significant gains in accuracy and AUC scores; nonetheless, AD123 presented a multi-class classification challenge. For example, the TabSeq model's accuracy improved from 66.67\% to 67.68\% for the AD123 target. Similarly, substantial improvements were noted for the ABETA12 target respectively 64.44\% to 75.13\% for TabSeq. By efficiently selecting and rearranging features according to their informative contribution to the predictive goal, feature ordering improved the performance of deep learning models. These results highlight the importance of adopting feature ordering to refine model predictions for complex datasets. The studies conducted with the WDBC dataset (Table \ref{tab:3ab}) demonstrate how feature ordering can improve model accuracy and AUC scores. The TabNet, NODE, and TabSeq versions performed better when feature ordering was used. Notably, compared to its excellent performance without feature ordering, the TabSeq model's accuracy increased, reaching 94.71\% with feature ordering. TabTransformer boosts AUC; TANGOS shows lower accuracy; TabPFN excels in feature-limited datasets like WDBC but without feature ordering. We employed k-means clustering with 5 clusters for the ADNI dataset and 3 for the WDBC dataset, ordering them ascendingly. As Fig. \ref{fig:cases} shows, increasing cluster numbers initially enhanced performance for most of the cases but eventually declined after specific points. These results affirm that our feature ordering approach boosts model accuracy and AUC across datasets, noting its importance in advancing deep learning for tabular data.
\begin{table}
  \caption{Comparative results on autoimmune diseases' dataset for different models (\# = without feature ordering, * = with feature ordering, Acc. = Accuracy).}
  \label{tab:1}
  \centering
  \begin{tabular}{|c|c|c|c|c|}
    \hline
    \textbf{Model} & \textbf{Acc.\#} & \textbf{AUC\#} & \textbf{Acc.*} & \textbf{AUC*} \\
    \hline
    \multicolumn{5}{|c|}{Case 1: H vs SLE+RA+SS+SV} \\
    \hline
    Linear SVM \cite{b37} & N/A & 0.94 & N/A & N/A\\
    \hline
    TabSeq (ours) & 85.42\% & 0.92 & 87.23\% & 0.92\\
    \hline
    LSTM & 82.11\% & 0.88 & 85.11\% & 0.91\\
    \hline
    DAE-LSTM & 75.79\% & 0.80 & 82.98\% & 0.80\\
    \hline
    DAE & 76.84\% & 0.80 & 80.85\% & 0.86\\
    \hline
    1D CNN & 72.92\% & 0.84 & 79.17\% & 0.44\\
    \hline
    TabNet \cite{b1} & 77.08\% & 0.85 & 83.33\% & 0.50\\
    \hline
    NODE \cite{b6} & 89.58\% & 0.92 & 20.83\% & 0.40\\
    \hline
    TabTransformer \cite{b2} & 82.98\% & 0.84 & 87.23\% & 0.75\\
    \hline
    TANGOS \cite{b68} & 82.98\% & 0.61 & 82.98\% & 0.65\\
    \hline
    \multicolumn{5}{|c|}{Case 2: SLE vs RA+SS+SV } \\
    \hline
    Linear SVM \cite{b37} & N/A & 0.96 & N/A & N/A\\
    \hline
    TabSeq (ours) & 86.11\% & 0.87 & 91.67\% & 0.96\\
    \hline
    LSTM & 81.94\% & 0.86 & 86.11\% & 0.93\\
    \hline
    DAE-LSTM & 79.17\% & 0.81 & 86.11\% & 0.87\\
    \hline
    DAE & 84.72\% & 0.84 & 86.11\% & 0.91\\
    \hline
    1D CNN & 80.56\% & 0.81 & 80.56\% & 0.31\\
    \hline
    TabNet \cite{b1} & 91.67\% & 0.60 & 94.44\% & 0.74\\
    \hline
    NODE \cite{b6} & 83.33\% & 0.82 & 80.56\% & 0.68\\
    \hline
    TabTransformer \cite{b2} & 86.11\% & 0.77 & 80.56\% & 0.54\\
    \hline
    TANGOS \cite{b68} & 86.12\% & 0.39 & 86.12\% & 0.73\\
    \hline
  \end{tabular}
\end{table}
\begin{table}
  \caption{Performance on ADNI and WDBC datasets.}
  \label{tab:3ab}
  \centering
  \begin{tabular}{|l|l|c|c|p{1.5cm}|p{1cm}|}
    \hline
    \textbf{Dataset} & \textbf{Model} & \textbf{Acc.\#} & \textbf{AUC\#} & \textbf{Acc.*} & \textbf{AUC*} \\
    \hline
    \multirow{6}{*}{ADNI} & \multicolumn{5}{c|}{\textbf{Target Attribute: AD123}} \\
    \cline{2-6}
    & TabNet \cite{b1} & 59.26\% & 0.68 & 66.67\% & 0.54 \\
    \cline{2-6}
    & NODE \cite{b6} & 59.26\% & 0.68 & 66.67\% & 0.54 \\
    \cline{2-6}
    & TabSeq (ours) & 66.67\% & 0.67 & 67.68\% & 0.61 \\
    \cline{2-6}
    & TabTransformer \cite{b2} & 66.67\% & 0.40 & 74.07\% & 0.70 \\
    \cline{2-6}
    & TANGOS \cite{b68} & 74.07\% & 0.70 & 74.07\% & 0.73 \\
    \cline{2-6}
    & \multicolumn{5}{c|}{\textbf{Target Attribute: ABETA12}} \\
    \cline{2-6}
    & TabNet \cite{b1} & 51.85\% & 0.52 & 74.07\% & 0.67 \\
    \cline{2-6}
    & NODE \cite{b6} & 59.26\% & 0.57 & 59.26\% & 0.68 \\
    \cline{2-6}
    & TabSeq (ours) & 64.44\% & 0.55 & 75.13\% & 0.71 \\
    \cline{2-6}
    & TabTransformer \cite{b2} & 59.26\% & 0.50 & 59.26\% & 0.50 \\
    \cline{2-6}
    & TANGOS \cite{b68} & 44.45\% & 0.50 & 44.45\% & 0.62 \\
    \hline
    WDBC & TabNet \cite{b1} & 91.86\% & 0.91 & 94.18\% & 0.99 \\
    \cline{2-6}
    & NODE \cite{b6} & 91.86\% & 0.99 & 93.71\% & 0.98 \\
    \cline{2-6}
    & TabSeq (ours) & 94.65\% & 0.91 & 94.71\% & 0.98 \\
    \cline{2-6}
    & TabTransformer \cite{b2} & 87.08\% & 0.91 & 60\% & 0.37 \\
    \cline{2-6}
    & TabPFN \cite{b69} & 94.19\% & 0.94 & 40.69\% & 0.50 \\
    \cline{2-6}
    & TANGOS \cite{b68} & 60\% & 0.76 & 60\% & 0.85 \\
    \hline
  \end{tabular}
\end{table}
\section{Conclusion}
This paper introduces TabSeq, a novel framework that employs feature ordering to enhance deep learning's performance on tabular datasets significantly. By integrating local ordering and global ordering within a DAE equipped with an MHA mechanism, our method systematically optimizes feature sequences to improve learning efficacy. Studies conducted on raw antibody microarray data and other medical datasets have underscored the capability of strategic feature sequencing to yield substantial performance gains. These empirical results reinforce feature ordering's potential as a game-changing technique in deep learning for tabular data. Its effectiveness is predominantly seen in sequence-based architectures, and its performance on low-dimensional datasets remains to be determined. Further research is needed to refine and extend the method's applicability to various architectures and datasets.

\vspace{-0.4cm}
\section*{Acknowledgement}
\vspace{-0.3cm}
This work was supported in part by grants from the US National Science Foundation (Award \#1920920, \#2125872, and \#2223793).

%
%
%
\bibliographystyle{splncs04}
\bibliography{main}

\end{document}